# Classifying logistic vehicles in cities using Deep learning

Salma Benslimane*,a, Simon Tamayoa, Arnaud de La Fortelle a

*aMines ParisTech – PSL Research University, Center for Robotics, Paris 75006, France*

## Abstract

Rapid growth in delivery and freight transportation is increasing in urban areas; as a result the use of delivery trucks and light commercial vehicles is evolving. Major cities can use traffic counting as a tool to monitor the presence of delivery vehicles in order to implement intelligent city planning measures. Classical methods for counting vehicles use mechanical, electromagnetic or pneumatic sensors, but these devices are costly, difficult to implement and only detect the presence of vehicles without giving information about their category, model or trajectory. This paper proposes a Deep Learning tool for classifying vehicles in a given image while considering different categories of logistic vehicles, namely: light-duty, medium-duty and heavy-duty vehicles. The proposed approach yields two main contributions: first we developed an architecture to create an annotated and balanced database of logistic vehicles, reducing manual annotation efforts. Second, we built a classifier that accurately classifies the logistic vehicles passing through a given road. The results of this work are: first, a database of 72 000 images for 4 vehicles classes; and second two retrained convolutional neural networks (InceptionV3 and MobileNetV2) capable of classifying vehicles with accuracies over 90%.

## 1. Introduction

In recent years urban freight transport has become more important for supporting a better life for people in urban areas (Taniguchi, 2015). A contemporaneous, relevant example is Internet shopping and home delivery: trends that simplify shopping experience for consumers but have a high impact on freight traffic. In fact, traditional distribution systems' retailers exchange large volume of goods via efficient and optimized utilization of trucks and heavy-duty vehicles, leaving last-mile transport to consumers. E-commerce however, relies on organizations characterized by reduced lead-times, which implies heavily substituting large freight transport by light-commercial vehicles LCVs causing an increase in last-mile delivery by vans and small trucks (Visser et al., 2014). Large trucks used for mass freight transportation are often held responsible for traffic blocking and increasing congestion (Dablanc, 2011). On the other hand, light commercial vehicles' growth raises concerns like environmental issues, traffic safety and congestion in residential areas.

Hence, a pressing need for appropriate guided regulation. Certainly, authorities could profit from knowing what vehicles occupy what roads at what times; such knowledge should be the starting point towards better regulation, and improvements on emissions, noise and road safety. Nonetheless, cities lack the instruments to quantify such transportation flows, they do not know what types of vehicles are occupying their road networks and as a result, they risk implementing regulation based on incomplete assumptions (Campbell, 1995).

The recent development in deep learning and computer vision in the fields of recognition and detection opens possibilities to developing smart cities and intelligent traffic systems while outperforming physical sensors.

---

\* Corresponding author. Tel.: +33 (0) 1 40 51 93 54
*E-mail address:* salma.benslimane@mines-paristech.fr

Research related to vehicle classification using neural networks gained growing attention in the past years as well as wide application scopes: model recognition (Munroe and Madden, 2005), traffic surveillance (Ozkurt and Camci, 2009) or even license plate location (Li et al., 2018). It is essential to highlight that there is an important parameter in this type of classification and that is the angle of view. In the past, some authors have proposed classification approaches for car recognition based on rear side view angles (Kafai and Bhanu, 2012), on one or two fixed cameras (Huttunen et al., 2016). However, there is a need for a classifier that generalizes to more angles of view.

In this paper we develop a classification network that would be applied to a wide range of angles of view. This implies that the proposed tool will be able to recognize vehicles in images coming from cameras with almost any possible orientation. A necessity for such classification algorithms is to have an appropriate database of vehicles on which to train. The work of (Yang et al., 2014) proposed large scale dataset for fine-grained categorization of model passenger car types, but today no large scale dataset contains specific logistic vehicles. As a result, one essential contribution of this research is developing a framework to create an annotated database of delivery vehicles images with varied angles of view.

We start with presenting the overall architecture of our dataset creation framework. We justify the categorization of logistic vehicles by size and functionality to 4 distinct classes (heavy, medium or light-duty trucks, passenger cars). Next we describe web scraping and processing steps to a clean and annotated logistic vehicles dataset. In the second part, we present the results of trained image classifiers: InceptionV3 and MobileNetV2 on the constructed dataset. We leverage previously trained networks on larger dataset with an approach called transfer learning and compare their performances.

As a result, given video feed or images of traffic, cities can count LCVs and trucks passing through a particular street or cross road. Cities would be able to have real knowledge of the road occupancy related to city logistics and thus improve their overall traffic management policies.

## 2. Dataset creation framework

Existing datasets, such as ImageNet include millions of images covering thousands of categories. It represents a large and eclectic resource of correctly annotated images covering numerous fields and subfields (Krizhevsky et al., 2012). A first intuitive idea would be to use ImageNet to construct a fine-grained image dataset of logistic vehicles images: taking images from existing subcategories and rearranging them to meet our needs into new ones. This approach has two major drawbacks: inaccurate categorization and imbalanced datasets.

In fact, subcategories of ImageNet do not align with the categorization of logistic vehicles we would like achieve. ImageNet was constructed based on WordNet (large lexical database of English words) developed by Princeton University, using the country's commonly used vehicles and appellations. Bearing in mind that trucks' norms and usage varying across countries, we cannot use ImageNet as a direct source for our dataset creation. As a result, we develop a method to build a dataset of fine-grained classes of our choice, meeting European specific norms (with the possibility to be generalized to other regions) and considering its most popular delivery vehicles.

Another important consideration that drove the framework creation is the number of images per category. In large datasets like ImageNet, images per category vary in number, which would result in imbalanced fine-grained datasets. Our framework creates balanced classes in size.

### 2.1. Proposed methodology

Creating datasets is a time-consuming task mostly due to annotation. Each image needs to be correctly labeled with its corresponding category.

To overcome the annotation burden and produce a balanced database, we suggest a different approach for database creation. The framework contains the three following blocks (Fig. 1):

- Categorization: define distinct categories of vehicles of which we want to construct the dataset with minimum overlap and list models and makes of each subcategories;
- Web Scraping: scrape images for the models of each subcategory from appropriate sources;
- Processing: clean and cleanse to delete corrupt images, then apply a classification network to crop the vehicles and delete inexact images not containing elements from the wanted subcategories.

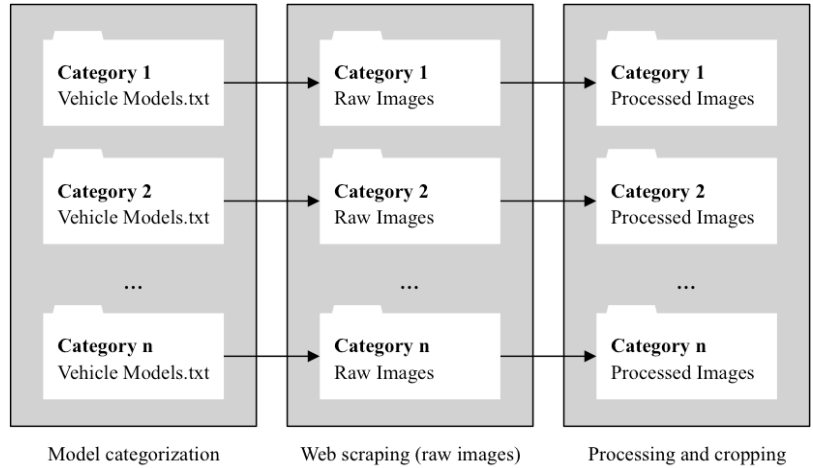

Fig. 1. Overview of the database creation framework.

*2.2. Vehicle categorization*

In order to create a database of subcategories with minimum overlap, we need to clearly define the classes' characteristics. Trucks have multiple dimensions and uses. Vans and minivans come in all shapes, sizes, forms and no conventional international categorization exists. To simplify, and mostly considering popular delivery vehicles in Europe, the categorization we used is as follows:

- Heavy-duty vehicles: trucks
- Medium-duty vehicles: vans
- Light-duty vehicles: minivans

Only using the previous keywords, overlap is inevitable between those categories and the distinction line is unclear. We refine the definition for each category by the ranges defined in Table 1.

Table 1. Vehicle categorization criteria.

| Type | Gross Vehicle Mass (in tons) | Total Height (in meters) |
|---|---|---|
| Light-duty vehicles | GVM ≤ 3.5 | H ≤ 2 |
| Medium-duty vehicles | GVM ≤ 3.5 | 2 < H ≤ 3 |
| Heavy-duty vehicles | GVM > 3.5 | H > 3 |

An example for each class is presented Fig 2.

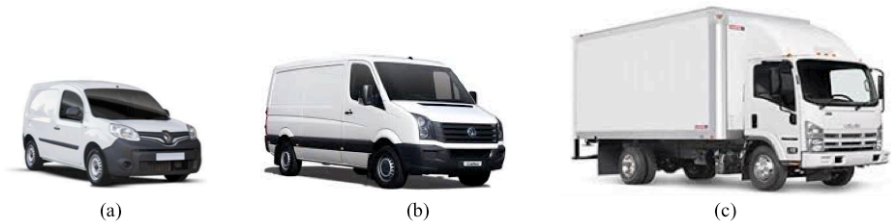

Fig. 2. Example of (a) light-duty (b) medium-duty and (c) heavy-duty vehicle.

Once the classes defined and given a vehicle's model and make, we can easily connect it to one of the previously discussed categories. Thus, we reverse engineer the process by creating for each class a list of car makes and model.

Assuming that the classification tool would be used to detect vehicles from streets images where other types of vehicles are to be detected and classified, we define another category containing other type of cars. This category can include for example: SUVs, sports cars, city cars, etc.

Note that as long as the categories chosen are distinct and have minimum resemblance or overlap, the framework can be generalized to other categories and domains.

## 2.3. Web Scraping

Instead of scraping images and then annotating each of them manually, we scape images by vehicle model to fall into a subcategory. Before going into further detail, it is important that we state that the scrapped images are only used to research purposes and that any commercial utilization is out of question. Two sources were used to obtain images:

- Web search engine such as Google, Bing, Yahoo, Baidu, etc.
- CAD model of vehicles

Table 2 presents examples of the vehicle models used in our categorization and querying:

Table 2. Vehicle categorization criteria.

| Light-duty vehicles | | Medium-duty vehicles | | Heavy-Duty vehicles | |
|---|---|---|---|---|---|
| Peugeot Expert | Nissan Caravan | Peugeot Boxer | Nissan NV400 | Mercedes Atego | Nissan Atleon |
| Renault Kangoo | Open Combi | Renault Master | Opel Movano | Renault D Wide | Kenworth K370 |
| Citroen Berlingo | Mercedes Vito | Citroen Jumper | Mercedes Sprinter | Volvo FH | Isuzu Serie N |
| Volkswagen Caddy | Fiat Scudo | Volkswagen Crafter | Fiat Ducato | MAN TGL | Scania R |
| Ford Transit Courier | … | Ford Transit 350 | … | Mitsubishi Canter | … |

Images from web search engines have the advantage of largely representing the same setting as for those from the streets. Meaning that most these images are taken with non-professional cameras, on random background, of real cars. Nevertheless, the drawback is that angles of view are non-variant: side and front views of vehicles are predominant. Only a few selection of images show the back angle of vehicles and fewer show aerial views.

Considering that most security cameras are recording from 3 to 4 meters above ground, views are not only various; they are heavily aerial. As a result, in an effort to consider most angles of view, we to complete our database with CAD images. We scrap from a professional CAD website high quality, high resolution, mono-color background, multi-angle images of vehicles. Although these pictures have minimal to non-existing noise and are not representative of the randomness of colors, backgrounds and low resolution of street images, they feed the database multi-angle view images. The resulting dataset is a mixture of pictures from web search engines and CAD website.

## 2.4. Image processing

Images scraped from the web, especially those from web search engines are not ready for immediate use. Some of them are corrupt, non-relevant, and/or noisy. Their sizes vary, backgrounds contain noise, as well as sometimes not containing vehicles or containing the interior of vehicles. If raw images are not processed, the network might learn useless features, and would perform poorly. Processing involves the two following steps:

- Deleting corrupt files;
- Passing images through the Yolo classification network (DarkNet) to detect vehicles and crop them.

Practically, we delete images that have not been downloaded properly, cannot be open or are corrupt. And we verify that image type and extensions in image name match. We then pass each scrapped image through pre-trained network Yolo (DarkNet-53) on ImageNet dataset, which has the advantage of being fast (78 FPS) as shown in Table 3.

Table 3. Comparison of characteristics of various networks(Redmon and Farhadi, 2018)

| Backbone | Top-1 | Top-5 | Bn Ops | BFLOP/2 | FPS |
|---|---|---|---|---|---|
| ResNet-101 | 77.1 | 93.7 | 19.7 | 1039 | 53 |
| ResNet-152 | 77.6 | 93.8 | 29.4 | 1090 | 37 |
| DarkNet-53 | 77.2 | 93.8 | 18.7 | 1457 | 78 |

Given an image, the network outputs object classes and bounding boxes. From all the detected objects, we select those labeled as vehicles and then crop them based on the bounding box (Fig. 3). We delete images containing other categories. As a result we obtain a dataset, containing predominantly centered images and we can have as many images of delivery vehicles as we want.

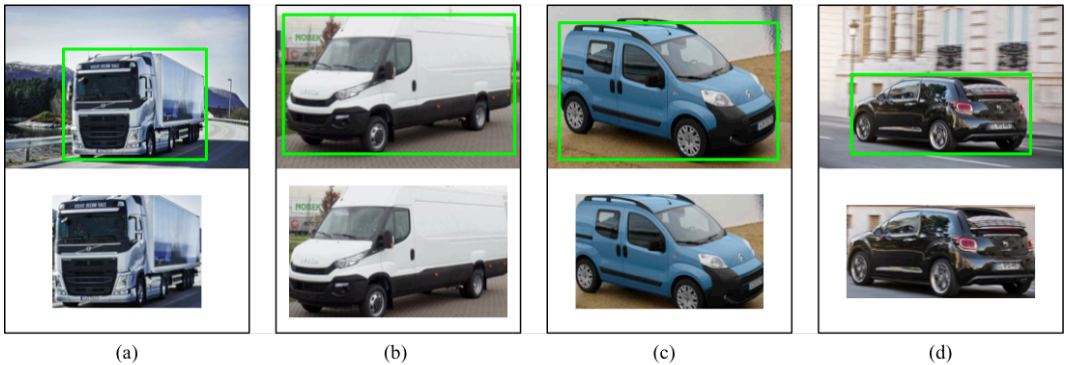

Fig. 3. (Top) Scrapped images of (a) heavy-duty, (b) medium-duty, (c) light-duty and (d) non-logistic vehicles passed through Yolo and their bounding box (in green) and (bottom) the cropped result.

## 2.5. Image processing

We apply the steps of the framework presented above, and we obtain a dataset of 72 000 images of 4 vehicles, with the following intermediate results:
- 4 chosen classes (Heavy, medium, light-duty vehicles and non-logistic)
- For each category, we query ~ 30 vehicle models (Table 2)
- For each category, we scrape ~ 20 000 raw images
- For each category, we obtain ~ 18 000 processed images

The dataset is mostly composed of centered images of vehicles covering most of the image area with minimum background. We notice that the dataset contains not only images of car exterior but also car interior that were classified by DarkNet as subcategories of car. Because the percentage of these images is marginal, it does not immediately negatively affect the training, however for a cleaner dataset a manual step of interior car images removal might be necessary.

## 3. Classifier training

Training a deep artificial neural network requires an important amount of data. Networks such as AlexNet, Inception of GoogleNet, or VGGNet for example are complex convolutional neural networks (CNN) with numerous layers and million parameters, needing large datasets for updating their weights (Russakovsky et al., 2015). So training complex networks from scratch to perform tasks such as classification entails having large datasets.

However, for multiple domains and applications, available datasets are restricted in size; making it difficult to train off-the-shelf complex deep networks with high performances. Transfer learning is an effective method used when confronted with a small dataset by leveraging pre-trained networks on large datasets like ImageNet (Shin et al., 2016). When small datasets contain images similar to those in existing large datasets, features learned on the large datasets can be transferred to learn on new images without having to start from scratch by initializing weights to random values.

These past years, new architectures for image classification are becoming deeper, more robust and outputting continuously improved results for image classification. Networks such as AlexNet, VGGNet or even GoogLeNet yielded high performance during the 2014 ILSVRC classification challenge. Inception, an architecture of GoogLeNet has relatively less parameters and lower computational cost compared to AlexNet and VGG while giving high classification accuracy (Szegedy et al., 2015).

On the other hand, complexity increases for those networks, which makes training long and lengthy. Networks, such as MobileNet developed for mobile applications are faster, less complex and have relatively high accuracies (even though smaller than more complex networks discussed previously (Howard et al., 2017)).

For training, we choose to compare two networks: InceptionV3 and MobileNetV2. For our application, considering that the constructed dataset contains approximately 72 000 images, we compare performances by applying transfer learning on both networks. Then we retrain the whole networks initializing the weight with pre-trained weights on ImageNet instead of random weights.

### 3.1. Classifier overview

The work of (Mensink et al., 2013) shows that generalizations of large dataset such as ImageNet-10K to unseen classes is possible. Consequently, we choose to extract features from the networks InceptionV3 and MobileNetV2 pre-trained on ImageNet as inputs for logistic vehicles' fine-grained classification.

Each image of the new dataset is passed through the network and features are stored. Features are outputs of the last convolutional layers of the CNN and inputs of the last fully connected layers. Once each image is passed through the network and its corresponding feature extracted, we train new the fully connected layers with the features as inputs and labels of the images as outputs. Processes summarized Fig 4.

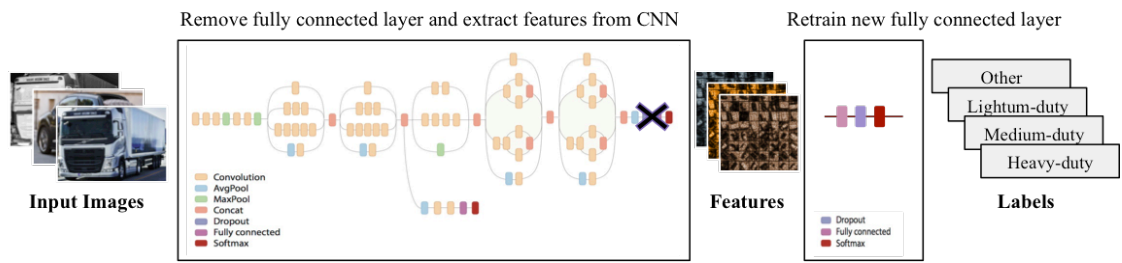

Fig. 4. Transfer learning process: passing images through convolutional layers of network to output features; then training new fully connected layers with the extracted features and the image labels.

### 3.2. Classifier performance

We train the two neural networks on 72 000 image dataset. We use a processor NVIDIA GeForce GTX 1080 TI. The accuracy of the methods is validated on 7 200 images test set (Table 4).

Table 4. Performance of transfer learning of CNNs on 10 epochs.

| Method | Model | Accuracy | Running time |
|---|---|---|---|
| Transfer Learning | InceptionV3 | $86.18 \pm 1.36\%$ | 194 s |
| | MobileNetV2 | $90.15 \pm 0.8\%$ | 65 s |
| Training network | InceptionV3 | $91.84 \pm 1.65\%$ | 60 min |
| | MobileNetV2 | $90.47 \pm 1.83\%$ | 44 min |

Retraining all the weights of the convolutional neural networks is costly in time and memory. Although transfer training gives slightly lower accuracy, its running time is considerably small compared to that of training the whole network.

To evaluate the classification networks on the created dataset of 4 classes, we compute the confusion matrix for MobileNetV2 network trained with transfer learning on the created dataset and tested on a sample set of 7200 images. The results of the confusion matrix are in Table 5.

Table 5. Normalized confusion matrix of MobileNet on a test sample of 7 200 images.

| | | Predicted label | | | |
|---|---|---|---|---|---|
| | | Non logistic | Medium-duty | Heavy-duty | Light-duty |
| True label | Non logistic | 94.11% | 0.86% | 0.62% | 4.40% |
| | Medium-duty | 1.99% | 88.07% | 3.01% | 6.93% |
| | Heavy-duty | 0.63% | 2.25% | 95.94% | 1.19% |
| | Light-duty | 6.42% | 7.97% | 0.91% | 84.71% |

MobileNetV2 classifies correctly vehicles into their categories with an overall accuracy of ~90 %. In details, it distinguishes heavy-duty vehicles clearly and best (95%) from other categories. As for the other categories, the network has more difficulty differentiating them: 6.42% and 7.97% of light-duty are classified as non-logistic and medium-duty respectively. 6.93% of medium-duty vehicles are classified as light-duty and 4.40% of non-logistic vehicles are classified as light-duty.

Manufacturers sometimes sell vehicles of different categories that have similar designs and shapes. Viewed from a distance and from a particular angle, the distinction is not trivial. Which is one of the main reasons causing confusing between light and medium-duty vehicles. Another consideration, is that the confusion between light-duty and non-logistic vehicles can be explained by the great similarity between some of these models, since many of them exist in 'passenger' and 'cargo' versions (ex. Renault Kangoo Fig 5), in these cases it is exactly the same vehicle and the only difference appears in the windows: the passenger model has glass and the model of load has metal sheet.

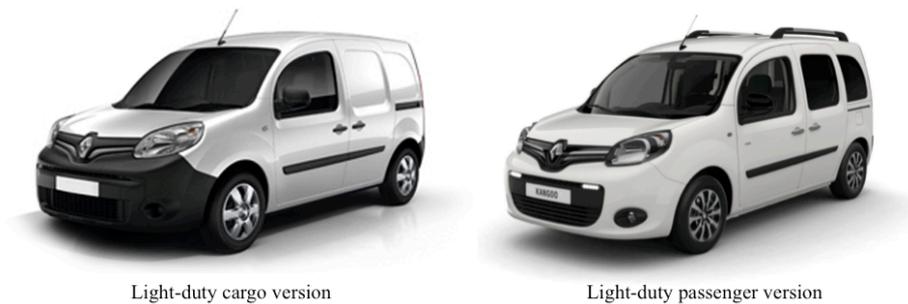

Light-duty cargo version          Light-duty passenger version

Fig. 5. Illustration of same model of 'Renault Kangoo' in cargo version and passenger version

## 4. Conclusion

This paper studied the problem of detecting and classifying logistic vehicles (vans, trucks, etc.) in images. These vehicles contribute significantly to traffic and emissions in cities. For this reason, it is important to define and quantify them in order to make better regulatory decisions.

The work presented here proposed an application of Deep Learning as a tool to detect and classify 3 types of logistics vehicles: light-duty, medium-duty, heavy-duty as well as non-logistic vehicles. More specifically, this research provides two contributions: (1) it proposes an architecture to create a database (of images) of logistic vehicles. And (2) it uses this base for the training of classification models using transfer learning.

A classification tool like the one we propose here is the starting point to create a tracking and counting system to monitor the flow of vehicles in cities. Indeed, cities have more and more recording cameras that could be used as "input" for this type of system. In fact, there are already commercial systems that use computer vision to count vehicles, but as of today, these tools are unable to distinguish between different types of logistic vehicles. This paper proposes elements to fill such a gap.

It is important to highlight the difficulties encountered in the creation of the database. Let's recall that the objective consisted in collecting an important quantity of labelled images of the different logistic vehicles. Additionally, the database should contain images taken from many points of view (i.e. several camera orientation angles). This generated two main difficulties. On the one hand we realized that the names of the types of vehicles are not homogeneous in different places of the world. For example in the USA the word "truck" corresponds to a 4x4 passenger vehicle, while in Europe it corresponds to a high-capacity freight transport vehicle. On the other hand, in the proposed architecture it is difficult to ensure that the base contains images of all viewing angles. In fact, it is rather unlikely for this to be the case. This is one of the limits of this research: it is possible that the classification models lose reliability when dealing with images taken from unusual points of view.

The results of this research are promising. A classification precision of approximately 90% is satisfactory for the purposes of classifying logistic vehicles in cities if we consider that today there are no other tools to do this type of counting. It is important to keep in mind that these results were obtained with a reasonably small image base for a Deep Learning project (72 000 images). We must then highlight that if we increase the base of images without increasing the classification categories, we will obtain even better results. Increase the size of the base can be done either by making more queries or by artificially modifying the images (rotating, zooming, changing colours, etc.). In both cases it is a simple procedure, but it should be noted that a larger base implies higher training times.

The classification results show that for the retrained networks it is very easy to distinguish heavy-duty vehicles. However, the analysis of the confusion matrix indicates that there is a significant difficulty in the distinction between medium-duty and light-duty vehicles since a significant proportion of medium-duty observations were classified as light-duty. This result is not surprising since in reality many of such models have quite similar features, if we consider, for example the case of a *Renault Kangoo* and a *Renault Master*, when we observe them from certain angles (from the front or from the top), even we are incapable of differentiating them. Another interesting confusion occurs between non-logistic vehicles and light-duty vehicles where some non-logistics observations were classified as light-duty because of model similarity. One of the perspectives of this research will be to increase the base of images with enough examples of 'passenger' and 'cargo' light-duty models in order to improve the classification scores in this type of vehicles.

## Acknowledgement

This research is part of the Urban Logistics Chair at MINES ParisTech PSL sponsored by *Marie de Paris* (Paris City Hall), *ADEME* (French Environment and Energy Management Agency), *La Poste*, *Pomona Group* and *Renault*.